**Algorithms used for the Cell Segmentation Benchmark Competition at ISBI 2019 by RWTH-GE**

**Authors**: Dennis Eschweiler[1], Johannes Stegmaier[1]

[1]Institute of Imaging and Computer Vision, RWTH Aachen University, Aachen, Germany.

**Email**: {dennis.eschweiler, johannes.stegmaier}@lfb.rwth-aachen.de

**Platform**: Windows 10 (x64)

**Prerequisites**: MATLAB 2018b (should be added to path, such that it can be opened from a command line with the command "matlab").

**Summary**

The presented algorithms for segmentation and tracking follow a 3-step approach where we detect, track and finally segment nuclei. In the preprocessing phase, we detect centroids of the cell nuclei using a convolutional neural network (CNN) for the 2D images and a Laplacian-of-Gaussian Scale Space Maximum Projection approach for the 3D data sets. Tracking was performed in a backwards fashion on the predicted seed points, *i.e.*, starting at the last frame and sequentially connecting corresponding objects until the first frame was reached. Correspondences were identified by propagating detections of a frame *t* to its preceding frame *t-1* and by combining redundant detections using a hierarchical clustering approach. The tracked centroids were then used as input to variants of the seeded watershed algorithm to obtain the final segmentation.

**Preprocessing**

*CNN-based 2D Centroid Detection*

Centroids of cell nuclei for the 2D data sets *Fluo-N2DH-GOWT1* and *Fluo-N2DL-HeLa* were identified using a U-Net [1]. To avoid that the network automatically learns to suppress detections in the border regions at this early stage, we cropped the original images to the field of interest specified in the annotation procedures (cropped border of *E=50* and *E=25* for the *Fluo-N2DH-GOWT1* and *Fluo-N2DL-HeLa* data sets, respectively). The intensity of the raw images was linearly scaled from the 0.4$^{th}$ and the 99.6$^{th}$ percentiles to the 8-bit value range [0, 255], while values above or below were set to the minimum and maximum value of this range. Moreover, the images were resized to 512x512 pixels for the CNN processing. We adapted the network described in our previous work by Eschweiler *et al.* [2] to work with single channel 2D images and additionally performed an intensity weighting of the loss, such that cells in low intensity areas contributed more to the loss. We trained a single detection network for both data sets using all available ground truth images. The obtained probability maps were then resized to original resolution, morphologically opened (*r=2*) and binarized using a threshold of *t=0.5*. Remaining connected seed points were separated using a watershed-based clump splitting applied on the Euclidean distance map of the inverted seed image (*i.e.*, intensity minima located in the nuclei centers). Detections touching the image border were removed from further processing.

*Laplacian-of-Gaussian Scale-Space Maximum Projection for 3D Centroid Detection*

For the 3D data sets *Fluo-N3DH-CE*, *Fluo-N3DL-TRIC* and *Fluo-N3DL-DRO* we used our previously published nucleus detection algorithm [3]. In brief, the original images were filtered using differently scaled Laplacian-of-Gaussian filters, where the selected scales were matched to the observed cell sizes (*Fluo-N3DH-CE*: $\sigma_{min} = 10, \sigma_{max} = 18$, *Fluo-N3DL-TRIC*: $\sigma_{min} = 4, \sigma_{max} = 6$, *Fluo-N3DL-DRO*: $\sigma_{min} = 4, \sigma_{max} = 4$). The 4D scale-space was reduced to a 3D image by a maximum intensity projection of the individual LoG-filtered images. Local maxima were then identified in the 3D LoG Scale-Space Maximum Projection (LoGSSMP) and we additionally allowed to detect intensity plateaus to reduce the number of false negative detections in cases where no single maximum pixel was present in the center of a nucleus. To reduce false positive detections in background regions, only detections with an intensity larger than the global mean plus two standard deviations of the LoGSSMP image intensities were considered. Moreover, the raw images were preprocessed using a 3D Gaussian filter (*Fluo-N3DH-CE*: $\sigma_{Gauss} = 3$, *Fluo-N3DL-TRIC*: $\sigma_{Gauss} = 3$), a 3D median filter ($r_{Median} = 1$, only first 50 frames of the *Fluo-N3DH-CE* data sets) or a grayscale 3D morphological erosion (*Fluo-N3DL-DRO*: $r_{Erode} = 2$). Redundant detections on intensity plateaus were combined in the subsequent tracking phase.

**Segmentation**

Segmentation was performed as the final step of the pipeline, see section *Post-Processing*.

**Tracking**

Tracking was performed in a backwards fashion by sequentially linking corresponding objects until the first frame was reached. Correspondences were identified by assigning a new tracking label to all unlabeled detections at frame *t* and by copying the labeled detections to the preceding frame *t-1*. At frame *t-1* we then performed a hierarchical clustering using Ward's linkage criterion and identify clusters that contain detections from time point *t* and *t-1*. The distance-based cut-off is determined based on the average spatial distance of each detection to a set of nearest neighbors (large in less dense frames, smaller in densely occupied frames). We empirically chose to use half the average distance to the 3$^{rd}$ to 8$^{th}$ nearest neighbors of each detection, *i.e.*, excluding potential redundant detections or division events from the computations. There are multiple cases of label presence in the clusters: (1) if a cluster contains no labeled detection, a new track label is introduced, (2) if one of the detections has a valid tracking label assigned, the label is copied for the cluster, (3) if two or more detections in the cluster have a valid tracking label assigned, a merge (*i.e.*, a cell division) is introduced by adding a new tracking label and by correctly assigning the predecessor and successor links, and (4) if only one labeled detection is contained, the track should end. The detections contained in a cluster are averaged and form the set of tracked objects at frame *t-1*. The steps above are repeated until the first frame of the sequence is reached.

In principle, this tracking approach is similar to a nearest neighbor tracking but additionally handles redundant detections by clustering nearby seed points. Moreover, cell divisions are naturally included in the algorithm design, as daughter cells are clustered together if their spatial distance falls below the cluster cut-off value. A drawback of the clustering-based approach, however, is the required global cut-off value that is used for the entire data set. This unavoidably leads to fragmented tracks if the data sets exhibit large density variations (*e.g., Fluo-N3DL-TRIC*). For the 3D data sets *Fluo-N3DL-TRIC* and *Fluo-N3DL-DRO* with varying densities and where only a subset of the cells should be tracked, we thus performed an additional nearest neighbor tracking in the forward direction. Selected tracks that ended earlier than the last frame were linked to their closest neighbors in subsequent frames if the respective detections were not yet occupied by another selected trajectory.

For the 2D data sets we additionally used the optical flow algorithm proposed by Farnebäck *et al.* [4] in its MATLAB implementation to propagate detections from one frame to another. This was performed in a forward and backward fashion, *i.e.*, detections of a frame *t* were copied to *t-1* and *t+1* and transformed using the velocity field proposed by the optical flow estimation step to be closer to the assumed true position. We found that this double-redundant seed propagation slightly improved the detection and tracking scores, potentially by eradicating flickering artifacts of the detection stage, where detections are sporadically missing in a few frames. Moreover, cell division events were more reliably detected with this approach as daughter cells were mostly correctly shifted on top of their mother cell, *i.e.*, ending up in the same cluster during the hierarchical clustering phase. For the 3D data sets we were not able to test this approach due to the limited amount of time and disabled the double-redundant seed detection as well as the optical flow predictions in the submitted version. However, we plan to investigate this possibility also for the 3D data sets in the future.

**Post-Processing**

The segmentation was largely based on a seeded watershed technique [5] with a few improvements to tune the results. The tracking labels were used to generate seed images with positions and labels identical to the tracked centroids. For the 2D data sets, we additionally generated a large background label based on a Euclidean distance map (EDM) of the seed points. The EDM was binarized using the maximally expected object radius as threshold value, such that all detected cells were surrounded by a background label with sufficient margin to allow for a proper segmentation. Moreover, seeds were dilated with the minimum expected object radius, to have a reasonable initialization for the seeded watershed and to prevent degenerate segmentations of only a few pixels. The edges identified by a Sobel edge detection were added to the inverted Gaussian-filtered raw images and then processed with a seeded watershed algorithm. Thus, the raw images were transformed such that intensity minima were located in the nuclei centers with additional boundaries provided by the Sobel operator to prevent leakage of background signal into the interior of the cells. While this approach worked properly for most nuclei, very dim regions were sometimes mis-segmented due to a very low signal to noise ratio. These issues could not be fixed before the result submission and need to be improved in future revisions. The background label as well as segments that clearly exceeded the maximum expected area were set to zero.

To obtain the instance segmentation for the 3D data sets, we used a slightly different processing pipeline. Based on the initial seed locations derived from the tracking phase, we analyzed small crops surrounding each of the detections in parallel fashion, similar to the TWANG segmentation approach described in [3]. Initially, we determine a binary threshold using Otsu's method separately for each of the cropped regions. However, instead of directly binarizing the small region with the identified threshold, we use a connected threshold image filter implemented in the Insight Toolkit (*itkConnectedThresholdImageFilter*). This filter is initialized with a seed point (the estimated centroid of the current nucleus) and then performs a region growing by extending to neighboring voxels if their intensity is above the threshold identified by the initial Otsu threshold. This step yields a single connected component that is connected to the seed point that was used for the initialization. As the binary connected component can potentially span over multiple neighboring nuclei, we perform an additional seeded watershed for clump-splitting on each of the regions by adding a seed in the center of the current nucleus of interest and by adding an artificial background marker on the image boundaries of the current region of interest. Thus, the center nucleus should be well separated from touching nuclei and background signal. The region size is set to the cut-off value used for the hierarchical clustering during tracking, *i.e.*, it is variable and depends on the density of the cells contained in each frame. The obtained segmentations for each individual cell are then combined to the resulting full resolution segmentation image. For the *Fluo-N3DL-TRIC* data set, we performed a maximum intensity projection of the 3D segmentation results and then resized the image in z to obtain the original number of slices. While this workaround does not necessarily provide good segmentation results in a biological sense, this was done to increase the chance of matching the slice-based annotations of the ground truth as 3D segments were systematically underestimated in the z direction and thus frequently ended prior to the slices of the ground truth annotations.

For both the 2D and the 3D data sets, the final segmentation images were double-checked with the tracking results and in cases where the segmentation algorithm erroneously missed a cell (*e.g.*, if the background label flooded a cell region), we manually added the detections again to provide segmentation images that are consistent with the tracking results.

**Implementation Details**

Data handling and tracking was implemented in MATLAB, the CNN-based seed detection was implemented in Python using Keras with the TensorFlow backend and compiled to a deployable executable using *pyinstaller*. The LoGSSMP seed detection and the segmentation pipelines were realized using the ITK-based C++ application XPIWIT [6].

**References**

1. Ronneberger, O., Fischer, P., & Brox, T. (2015). U-net: Convolutional Networks for Biomedical Image Segmentation. In International Conference on Medical Image Computing and Computer-Assisted Intervention (pp. 234-241). Springer, Cham.
2. Eschweiler, D., Spina, T. V., Choudhury, R. C., Meyerowitz, E., Cunha, A., & Stegmaier, J. (2018). CNN-based Preprocessing to Optimize Watershed-based Cell Segmentation in 3D Confocal Microscopy Images. arXiv preprint arXiv:1810.06933.
3. Stegmaier, J., Otte, J. C., Kobitski, A., Bartschat, A., Garcia, A., Nienhaus, G. U., Strähle, U. & Mikut, R. (2014). Fast Segmentation of Stained Nuclei in Terabyte-Scale, Time Resolved 3D Microscopy Image Stacks. PLOS ONE, 9(2), e90036.
4. Farnebäck, G. (2003). Two-Frame Motion Estimation Based on Polynomial Expansion. In Scandinavian Conference on Image Analysis (pp. 363-370). Springer, Berlin, Heidelberg.
5. Beare, R., & Lehmann, G. (2006). The Watershed Transform in ITK – Discussion and New Developments. The Insight Journal, 6.
6. Bartschat, A., Hübner, E., Reischl, M., Mikut, R., & Stegmaier, J. (2015). XPIWIT – An XML Pipeline Wrapper for the Insight Toolkit. Bioinformatics, 32(2), 315-317.